\documentclass[letterpaper, 10 pt, conference]{ieeeconf}  

\IEEEoverridecommandlockouts                              

\usepackage[noadjust]{cite} 
\usepackage[T1]{fontenc}
\usepackage[utf8x]{inputenc}
\usepackage{amssymb}
\usepackage{xcolor}
\usepackage{comment}
\let\labelindent\relax
\usepackage{enumitem}
\usepackage{url}
\usepackage{booktabs}
\usepackage{bbding}
\usepackage{pifont}
\usepackage{utfsym}
\usepackage{fontawesome}
\usepackage{multirow}
\usepackage{arydshln}
\usepackage{amsmath} 
\usepackage{soul}

\title{\LARGE \bf
Physics-Informed Machine Learning with Adaptive Grids for Optical Microrobot Depth Estimation}

\author{Lan Wei, Lou Genoud and Dandan Zhang
\thanks{
Lan Wei and Dandan Zhang are with the Imperial-X Initiative, Department of Bioengineering, Imperial College London, London, United Kingdom.
Lou Genoud are with the Department of Surgery and Cancer, Imperial College London, London, United Kingdom.
Corresponding: Dandan Zhang, {\tt\small d.zhang17@imperial.ac.uk}}
}

\begin{document}

\maketitle
\thispagestyle{empty}
\pagestyle{empty}

\begin{abstract}

Optical microrobots actuated by optical tweezers (OT) offer great potential for biomedical applications such as cell manipulation and microscale assembly.
These tasks demand accurate three-dimensional perception to ensure precise control in complex and dynamic biological environments.
However, the transparent nature of microrobots and low-contrast microscopic imaging challenge conventional deep learning methods, which also require large annotated datasets that are costly to obtain. 
To address these challenges, we propose a physics-informed, data-efficient framework for depth estimation of optical microrobots.
Our method augments convolutional feature extraction with physics-based focus metrics, such as entropy, Laplacian of Gaussian, and gradient sharpness, calculated using an adaptive grid strategy.
This approach allocates finer grids over microrobot regions and coarser grids over background areas, enhancing depth sensitivity while reducing computational complexity.
We evaluate our framework on multiple microrobot types and demonstrate significant improvements over baseline models.
Specifically, our approach reduces mean squared error (MSE) by over 60\% and improves the coefficient of determination ($R^2$) across all test cases. 
Notably, even when trained on only 20\% of the available data, our model outperforms ResNet50 trained on the full dataset, highlighting its robustness under limited data conditions. Our code is available at: https://github.com/LannWei/CBS2025.
\end{abstract}

\section{Introduction}
Microrobotic systems have attracted growing attention due to their diverse applications in biomedical applications~\cite{zhang2023advanced,lu2000fast,lin2024magnetic}. 
Advances in 3D microfabrication and materials science have enabled the design, fabrication, and functionalization of microrobots with complex, task-specific geometries~\cite{zhang2022fabrication,li20213d,rayabphand2025magni}.
Optical tweezers (OT) have emerged as a promising technique in micromanipulation and can actuate microrobots for indirect manipulation of biological targets such as cells~\cite{thakur2014indirect}. 
In most OT platforms, microrobots are monitored using a monocular microscope equipped with a CCD camera.

However, the transparent nature of microrobots, combined with the limitations of parallel projection in microscopy~\cite{boyde1985stereoscopic}, poses significant challenges for three-dimensional (3D) localisation. 
In particular, movements along the optical (z) axis are not readily captured, making it difficult to infer depth from a single image~\cite{chang2019deep,zhang2020distributed}.
Depth estimation refers to the process of determining the vertical distance between the microrobot and the microscope’s focal plane.
Reliable depth estimation is essential for closed-loop control, autofocusing, and intuitive human-robot interaction. It allows the system to maintain the microrobot in focus, perform precise navigation, and avoid collisions or damage to surrounding biological structures.


\begin{figure}[!t]
\centering
\includegraphics[width=1\hsize]{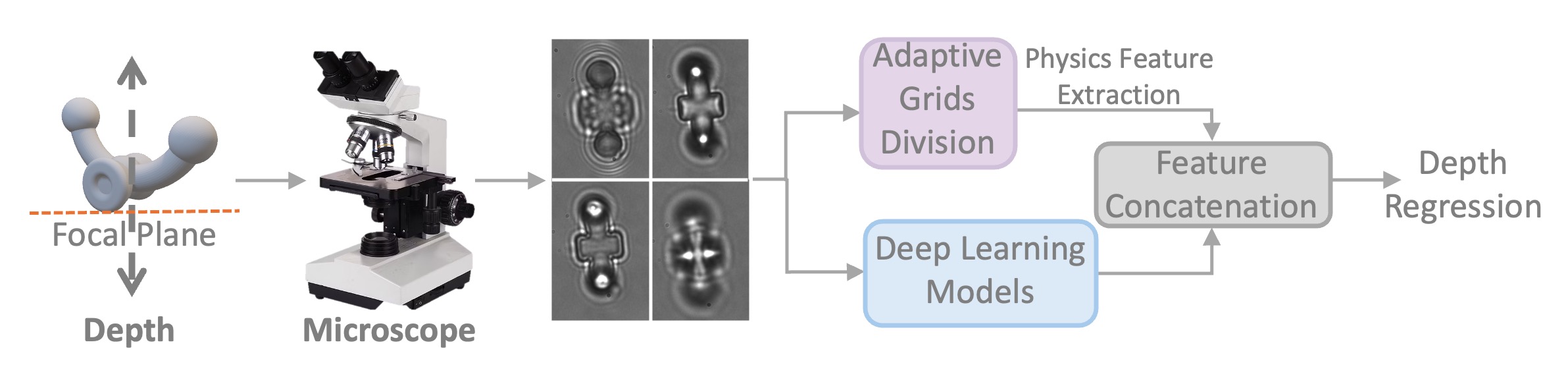}
\vspace{-0.8cm}
\caption{
Overview of the proposed physics-informed depth estimation framework. 
Depth is defined as the microrobot’s displacement relative to the microscope’s focal plane. 
Microscopic images are processed via adaptive grid-based feature extraction and deep learning for depth prediction.
}
\label{fig-concept}
\vspace{-0.6cm}
\end{figure}%


Traditional techniques for estimating the depth of microobjects have been studied. 
For instance, feature-based vision methods have been applied to estimate the depth and orientation of micro-grippers, nanorobots, and carbon nanotubes using scanning electron microscopy (SEM)~\cite{marturi2018image,kudryavtsev2017full,marturi2016visual}. 
However, these approaches are not readily applicable to optical microrobots, particularly those with high transparency. 
The rapid changes in depth during manipulation, coupled with the difficulty of extracting distinct features from transparent objects, present significant challenges.

In recent years, deep learning-based methods have been introduced to enhance the sensing capabilities of microrobots. 
By leveraging end-to-end supervised learning, neural networks have demonstrated better performance in mapping raw image data to target depth values compared to traditional focal measurement techniques. 
Grammatikopoulou et.al combined a convolutional neural network (CNN) with Long Short-Term Memory (LSTM) to estimate the 3D pose and depth of transparent microrobots observed under OT~\cite{grammatikopoulou2017depth,grammatikopoulou2019three}. 
Zhang et.al. proposed a hybrid approach combining Gaussian Process Regression (GPR) with a ResNet architecture for pose and depth estimation, integrating the strengths of focus measurements and deep residual networks~\cite{zhang2020data}.
Despite these advances, most deep learning-based methods require large annotated datasets, which are costly and labour-intensive to acquire in microrobotic applications. 
Moreover, existing models often fail to fully exploit the available physical information in images, and their estimation accuracy remains limited, especially when only small datasets are available.

\begin{figure}[!t]
\centering
\includegraphics[width=1\hsize]{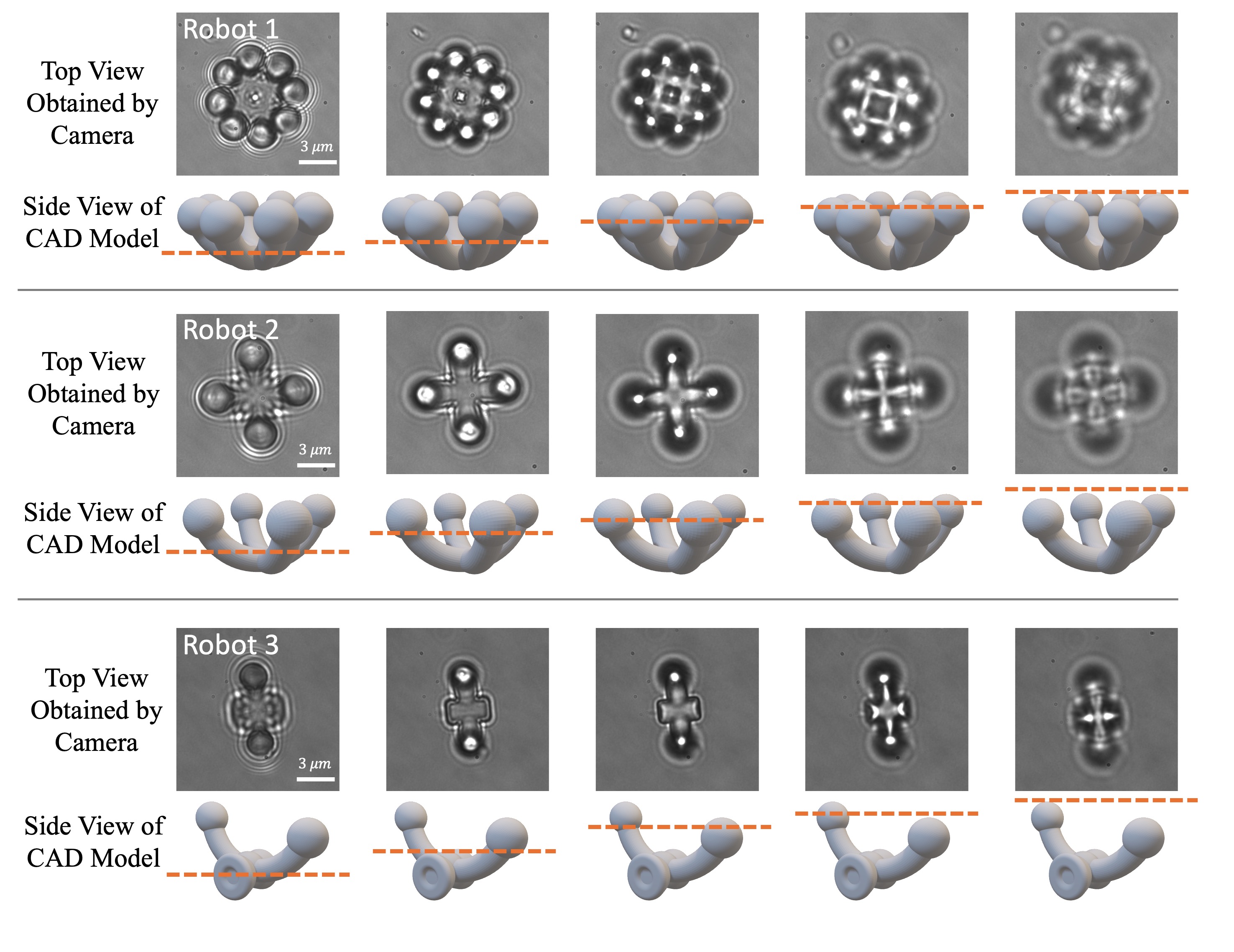}
\vspace{-0.9cm}
\caption{Visualization of microrobots at varying depths relative to the microscope focal plane (dashed line). Each row shows the camera top view and the corresponding CAD-rendered side view.
}
\label{fig-robots}
\vspace{-0.6cm}
\end{figure}%

Physics-informed machine learning has emerged as a promising paradigm to address these challenges by embedding prior physical knowledge into learning processes, thereby reducing the reliance on large datasets~\cite{xu2023physics}.
Motivated by this idea, we propose a physics-informed depth estimation framework for optical microrobots that explicitly integrates physical focus cues, such as image entropy, Laplacian of Gaussian, and gradient-based sharpness metrics, into the deep learning pipeline (as shown in Fig.~\ref{fig-concept}). 
These handcrafted features capture depth-dependent visual patterns that are difficult to learn directly from limited training data, and thus provide valuable information to standard visual features extracted by CNNs.

In addition, conventional CNNs typically apply fixed-size convolutional kernels uniformly across the entire image~\cite{zhang2017scale}, without considering the spatial salience of different regions. 
However, in microscopic imaging of microrobots, depth-relevant information is concentrated within the microrobot itself, while the surrounding background often contains redundant or non-informative content. 
To address this issue, we introduce an adaptive grid-based feature extraction strategy that dynamically allocates finer grids around the microrobot and coarser grids in the background. This approach not only enhances the sensitivity to local structural variations but also reduces computational redundancy.

The extracted physics features from each adaptive grid are aggregated and concatenated with the deep visual features from the neural network. 
This fusion takes place prior to the final regression layer, enabling the network to jointly leverage both domain knowledge and learned representations. 
Our proposed design functions as a plug-and-play module that can be seamlessly integrated into a variety of network architectures. 
In this work, we adopt ResNet-50~\cite{he2016deep} as the backbone to validate the effectiveness of the approach.
Overall, the proposed framework improves both the accuracy and data efficiency of depth estimation. 
It is particularly well-suited for microrobot perception tasks in data-scarce, noisy, and high-precision environments such as those found in OT-based biomedical platforms.

\section{Methodology}
\begin{figure*}[!t]
\centering
\includegraphics[width=0.9\hsize]{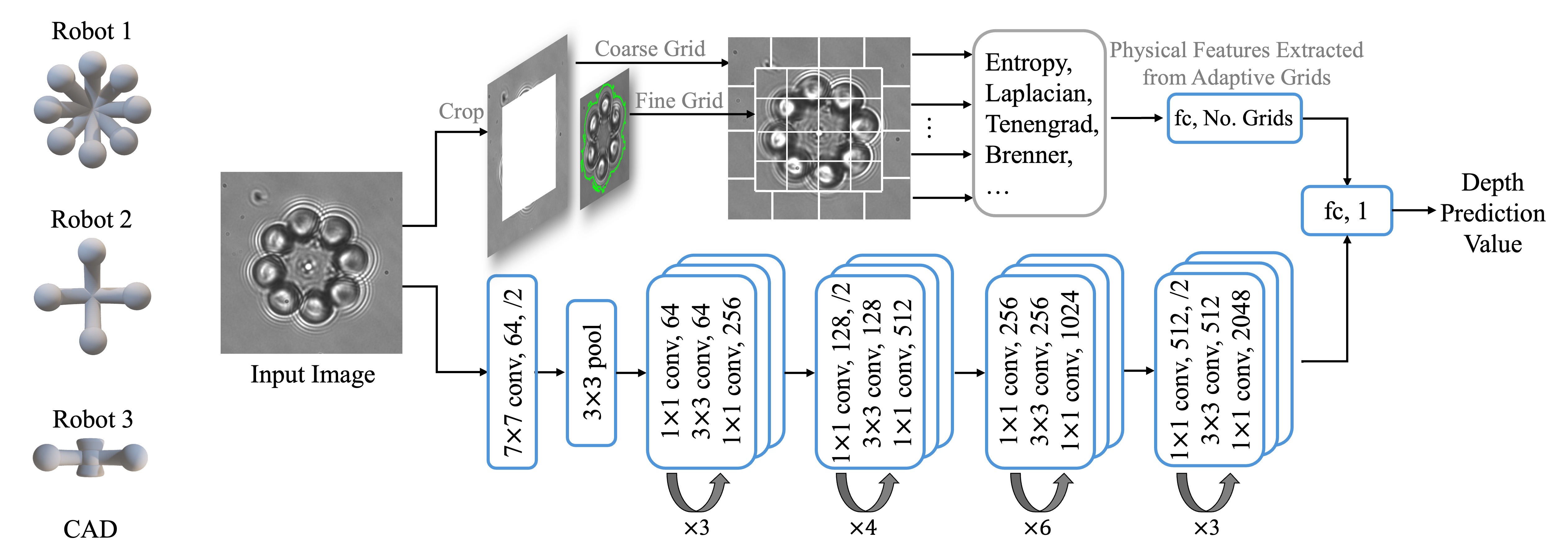}
\vspace{-0.5cm}
\caption{Architecture of the proposed physics-informed depth regression framework. The input image is processed in two parallel streams: one through a ResNet-50 backbone for deep feature extraction, and the other through adaptive grid partitioning for physics-based feature extraction (e.g., entropy, Laplacian, Tenengrad, Brenner). The two feature sets are concatenated before the final fully connected layer to produce the depth prediction value.
The left is the CAD view of the three robots we use.
}
\label{fig-framework}
\vspace{-0.6cm}
\end{figure*}%

In this study, we investigate three types of transparent microrobots to evaluate the proposed depth perception framework. 
The primary objective is to estimate the vertical position of the microrobot relative to the microscope’s focal plane, which is formulated as a supervised regression problem. 
As illustrated in Fig.~\ref{fig-robots}, vertical displacements are generated by actuating a piezoelectric stage that moves the microrobot along the optical (z) axis. 
Accurate prediction of this displacement is essential for achieving precise 3D localisation, which in turn enables closed-loop control and high-precision micromanipulation.
The architecture of the proposed physics-informed depth regression framework are shown in Fig.~\ref{fig-framework}.

\subsection{Physics-Based Feature Extraction}
As a microrobot undergoes depth variation under a microscope, its visual appearance, such as sharpness, contrast, and edge clarity, exhibits noticeable changes. These variations can be leveraged to infer depth by extracting physical features from monocular microscope images. 
However, relying on a single focus-related metric is often insufficient for accurate depth reconstruction, as different depths may yield similar values for a given feature due to noise or local structural ambiguity. Therefore, it is necessary to combine multiple complementary focus metrics that capture distinct visual cues associated with depth changes. Here, we construct a feature set composed of six focus measurement operators that encode different aspects of image sharpness and structural information:


\begin{enumerate}[nosep,leftmargin=*]
\item \textbf{Entropy}, which quantifies the complexity or randomness of pixel intensity distributions, with higher values typically indicating richer image content at in-focus depths~\cite{mello2019entropy};
\[ 
\text{M}_1 = - \sum_{i=0}^{L-1} p_i \log_2(p_i), 
\]
where \( p_i \) is the probability of intensity level \( i \), and \( L \) is the number of gray levels.
\item \textbf{Laplacian of Gaussian (LoG)}, which captures edge intensity changes through second-order derivatives, reflecting the level of detail present in the image~\cite{kong2013generalized};
\[
\text{M}_2 = \nabla^2 (G(x, y, \sigma) * I(x, y)),
\]
where \( G(x, y, \sigma) \) is a Gaussian kernel, \( * \) denotes convolution, and \( \nabla^2 \) is the Laplacian operator.
\item \textbf{Tenengrad}, which evaluates gradient magnitude over the entire image, measuring edge clarity and focus sharpness~\cite{srinivasan2018comparative};
\[
\text{M}_3 = \sum_{x, y} G_x(x, y)^2 + G_y(x, y)^2,
\]
where \( G_x \) and \( G_y \) are the horizontal and vertical Sobel gradients, respectively.
\item \textbf{Brenner measure}, which computes the squared difference between adjacent pixels along rows, assuming that focused images exhibit stronger local intensity variations~\cite{brenner2005building};
\[
\text{M}_4 = \sum_{x, y} \left(I(x + 2, y) - I(x, y)\right)^2.
\]
\item \textbf{Gray Level Variance}, which measures the statistical variance of pixel values across the image, based on the principle that focused images tend to have more spread in intensities~\cite{panda1978image};
\[
\text{M}_5 = \frac{1}{N} \sum_{x, y} \left(I(x, y) - \mu\right)^2,
\]
where \( \mu \) is the mean pixel intensity and \( N \) is the total number of pixels.
\item \textbf{Maximum Absolute Gradient}, which captures the strongest single-edge response in the image and serves as an indicator of sharpness in high-contrast regions.
\[
\text{M}_6 = \max_{x, y} \left| \nabla I(x, y) \right|, \text{where } \nabla I(x, y) = \sqrt{G_x^2 + G_y^2}.
\]
\end{enumerate}

By concatenating these features, we aim to build a reliable and discriminative representation of depth-dependent image characteristics. 
These physics-informed descriptors are extracted locally from adaptively segmented image patches (as detailed in the next section) and subsequently fused with deep visual features for robust depth regression.

\begin{table*}[!t]
\centering 
\caption{Quantitative comparison of depth regression performance across three types of microrobots. 
}
\vspace{-0.2cm}
\begin{tabular}{c|cc:cc:cc:ccc}
\hline \hline 
\multirow{2}{*}{Method} & \multicolumn{2}{|c:}{Robot 1} & \multicolumn{2}{c:}{Robot 2} & \multicolumn{2}{c:}{Robot 3} & \multicolumn{3}{c}{Model Size} \\
\cline{2-10}
& MSE($\downarrow$) & $R^2$($\uparrow$) & MSE($\downarrow$) & $R^2$($\uparrow$)  & MSE($\downarrow$) & $R^2$($\uparrow$)  & Params (MB) & GFLOPs & Throughput (img/s)\\ \hline 
CNN      & 0.479 & 0.965& 0.117&0.995 & 0.064& 0.990& 25.78& \textbf{0.53} & \textbf{261726.53}\\
ResNet18 & 0.082 & 0.994& 0.099& 0.996& 0.060&0.990 & \textbf{11.18}& 1.82 & 56967.67\\
ResNet50 & 0.075 & 0.994 & 0.109& 0.995& 0.059& 0.991&23.51 &4.13 & 23028.00\\ 
Ours &\textbf{0.027} & \textbf{0.999}& \textbf{0.082}& \textbf{0.997}& \textbf{0.054}& \textbf{0.992} & 24.56 & 4.13 & 22243.76\\   
\hline \hline 
\end{tabular}
\\MSE values are reported in micrometre squared (\(\mu\mathrm{m}^2\)).
The best result in each column is shown in \textbf{bold}.
\\Arrows indicate metric direction: \(\downarrow\) lower is better; \(\uparrow\) higher is better.
\label{table-depth_result}
\vspace{-0.2cm}
\end{table*}

\subsection{Adaptive Grid Construction}
To enhance computational efficiency while preserving informative spatial structure, we introduce an adaptive grid construction strategy for localised physical feature extraction. 
Rather than applying uniform grids across the entire image, our method adaptively partitions the input based on the spatial distribution of the microrobot.

Each input image, sized 224 × 224 pixels, is first analysed to identify the microrobot’s location. 
We employ contour detection based on Gaussian blurring~\cite{gedraite2011investigation} and Otsu thresholding~\cite{feng2017multi} to segment the microrobot from the background, as shown in the upper part of Fig.~\ref{fig-framework}. 
A bounding box is then computed around the largest detected contour, which is assumed to correspond to the microrobot. 
To include contextual information, the bounding box is expanded by a fixed ratio (we use 1.2 in our implementation).
The image is then divided into two spatial regions: a foreground region centred on the microrobot and a background region covering the rest of the image. 
The foreground region is partitioned into a finer grid of size 6×6, corresponding to local patches of approximately 37 × 37 pixels. 
These patches capture subtle variations in sharpness and structure caused by depth changes. 
The background region is coarsely divided into a 4×4 grid (patch size ~56 × 56 pixels), as it contains minimal task-relevant information.

Physical features such as entropy, Laplacian, and gradient-based metrics are computed independently for each grid cell. 
To ensure consistent input size for the model, we maintain a fixed number of patches (6×6 + 4×4 = 52) by zero-padding. 
Each patch contributes a vector of physics-based descriptors, which are concatenated to form a global physics feature vector.
Rather than relying on traditional mapping functions such as nonparametric Bayesian models to regress depth from physical features only, we adopt a more unified strategy. 
The global physics feature vector is concatenated with deep visual features extracted from the ResNet-50 backbone~\cite{he2016deep}, just before the final fully connected regression layer. 
This fusion preserves the expressive power of deep convolutional networks while incorporating physics-informed priors that enhance the model’s sensitivity to depth-induced visual changes.

The proposed methodology integrates physics-based focus cues and adaptive spatial reasoning to enable accurate and efficient depth estimation of transparent microrobots. 
Physical features extracted from adaptive grid patches provide complementary information to high-level visual representations, especially under challenging visual conditions. 
By tailoring the granularity of analysis to the spatial salience of the microrobot, the framework reduces computational redundancy. 
Moreover, the modular design of our physics-informed feature extraction module allows it to be seamlessly integrated with various deep learning architectures, demonstrating strong potential for broader application in microrobotic perception tasks.

\section{Experiments}

\begin{figure*}[!t]
\centering
\includegraphics[width=0.85\hsize]{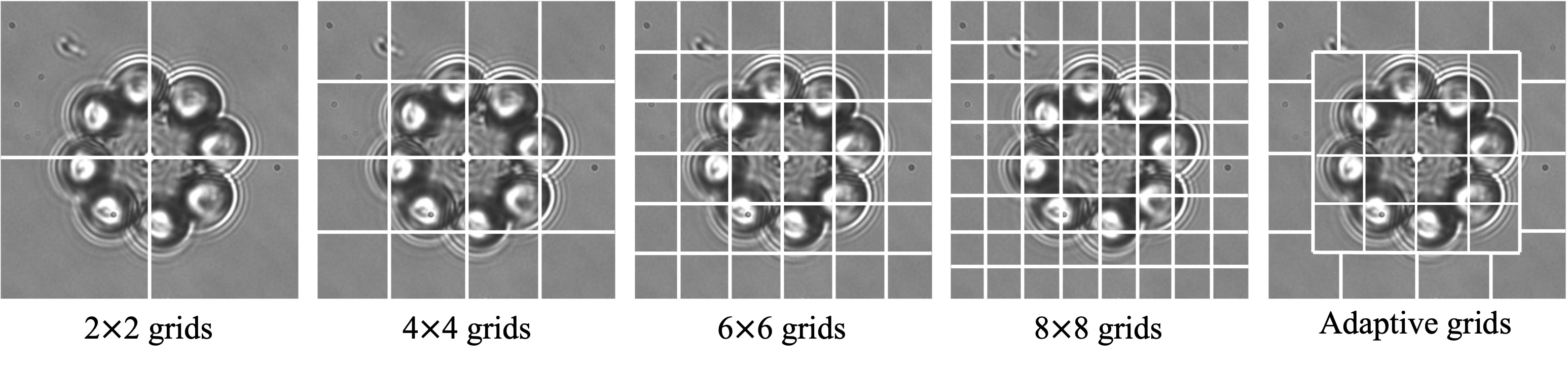}
\vspace{-0.5cm}
\caption{Visual comparison of uniform and adaptive grid strategies. While uniform grids divide the entire image equally, the adaptive grid method allocates finer grids to the microrobot region and coarser grids to the background.}
\label{fig-grids}
\vspace{-0.5cm}
\end{figure*}%

\subsection{Experimental Setup}
All experiments were conducted using PyTorch 1.12 in a Python 3.8 environment. Model training and evaluation were performed on a workstation equipped with a single NVIDIA A100 GPU featuring 80 GB of memory. The CUDA runtime version used was 12.0. For throughput evaluation, inference speed was measured on the A100 GPU with a batch size of 128, utilising 32-bit floating point (FP32) precision.

For the model backbone, we employed ResNet-50~\cite{he2016deep} initialised with pre-trained weights from the \textit{torchvision} library~\cite{torchvision2016}. 
The network was optimised using the Adam optimiser with an initial learning rate set to 0.001, and each model is fine-tuned for 20 epochs.

\subsection{Data Collection}
Microrobots were fabricated in IP-L photoresist (Nanoscribe GmbH, Germany) using a Nanoscribe 3D printer. Experiments were conducted on an OT system (Elliot Scientific, UK) integrated with a nanopositioner (Mad City Labs Inc., USA), and a CCD-coupled microscope captured images of the microrobots. During acquisition, the microrobot was stepped through axial depths and out-of-plane orientations so that each frame corresponded to a specific depth–pose pair~\cite{zhang2022micro}. The recorded images had dimensions of \(678\times 488\) pixels, and more than 5000 frames were collected for depth-estimation experiments; additional details of the data-collection procedure are provided in~\cite{wei2025dataset}. 

For preprocessing, raw images were converted to grayscale and denoised with a pixel-wise adaptive low-pass Wiener filter~\cite{lim1990two}. Binarisation with an automatically determined threshold was followed by Canny edge extraction~\cite{canny1986computational}. The microrobot centroid was then localised, and a \(256\times 256\) region of interest (ROI) centred on the microrobot was cropped for subsequent analysis.


\subsection{Evaluation Metrics}
To comprehensively evaluate the model's performance on depth regression tasks,
we use the following evaluation metrics:
Mean Squared Error (MSE): the average squared difference between the predicted and true depth values, where larger deviations are penalised more heavily;
Coefficient of Determination ($R^2$): a statistical measure that indicates how well the predicted values approximate the ground truth data variance, with an $R^2$ value closer to 1 representing better model performance.

Beyond predictive accuracy, we also assess the models from a computational efficiency perspective. Specifically, we report:
Model size, measured by the number of parameters (in megabytes, MB);
Computational complexity, quantified by the number of floating point operations per second (FLOPs), expressed in gigaflops (GFLOPs);
Real-time inference throughput is defined as the number of images processed per second during model inference.
These additional metrics enable a detailed evaluation of the trade-off between model accuracy and computational efficiency, which is particularly important for real-time microrobotic applications.

\subsection{Depth Regression}

As shown in Table~\ref{table-depth_result}, our proposed method consistently outperforms baseline models across all three microrobots in terms of both MSE and $R^2$. Specifically, our approach achieves the lowest MSE and the highest $R^2$ on Robot 1 (0.027 $\mu m$, 0.999), Robot 2 (0.082 $\mu m$, 0.997), and Robot 3 (0.054 $\mu m$, 0.992). 
Compared to baselines, our model achieves higher accuracy while maintaining comparable model size (24.56 MB) and computational cost (4.13 GFLOPs). 
Although throughput is slightly lower than that of smaller models like ResNet50, the gain in regression precision demonstrates the effectiveness of incorporating physics-informed features and adaptive grid division for microrobot depth perception.

\subsection{Ablation Studies}
\begin{figure}[!t]
\centering
\includegraphics[width=1.0\hsize]{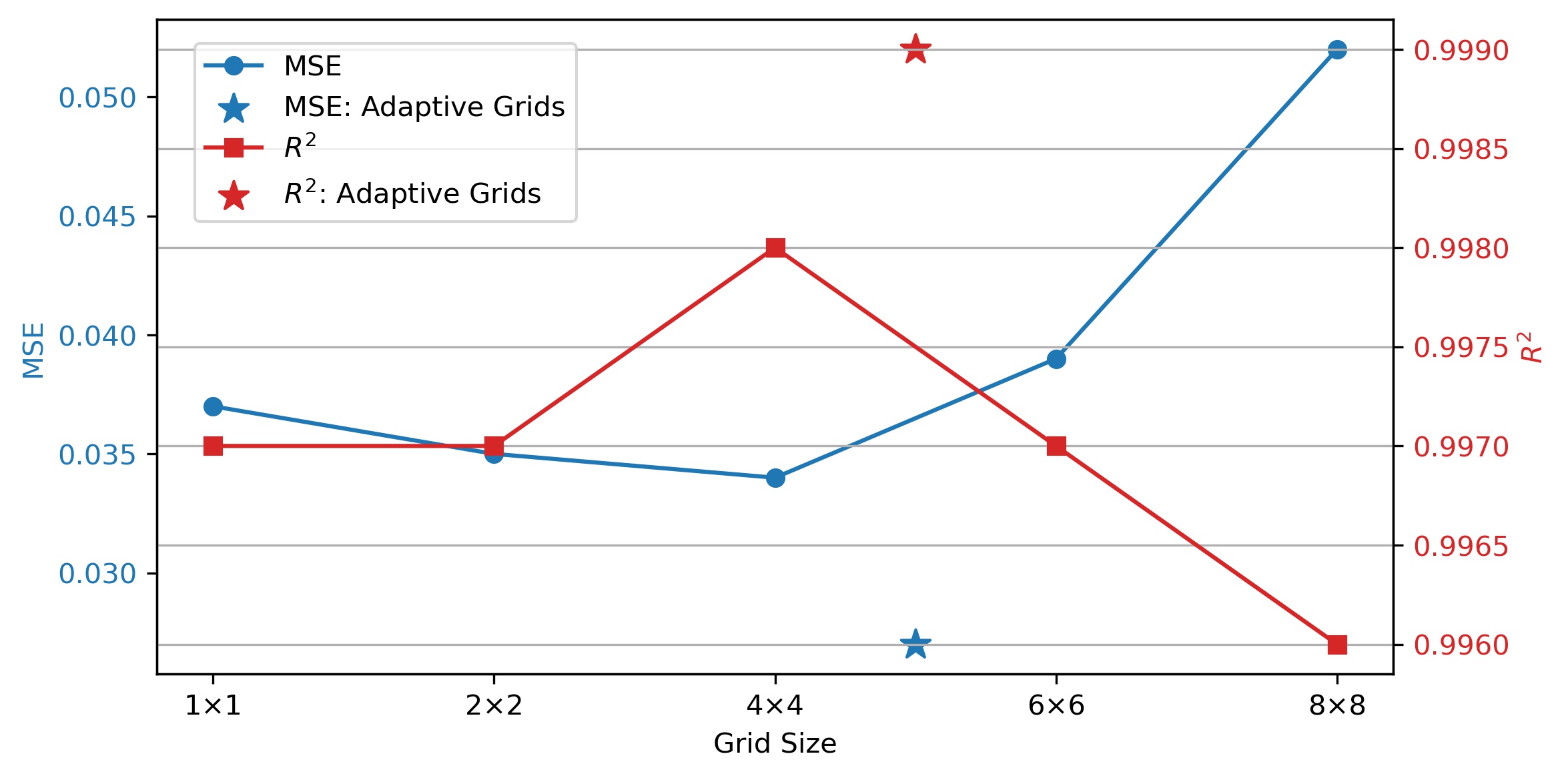}
\vspace{-0.8cm}
\caption{Ablation study results on robot 1 depth estimation using different grid strategies. The adaptive grid approach achieves the lowest MSE and the highest $R^2$, outperforming all uniform grid configurations.}
\label{fig-ablation}
\vspace{-0.2cm}
\end{figure}%

To investigate the impact of spatial resolution in physical feature extraction, we conducted an ablation study by dividing the input image into grids of varying granularity. 
As illustrated in Fig.~\ref{fig-grids}, we compared four uniform grid strategies—2$\times$2, 4$\times$4, 6$\times$6, and 8$\times$8—with our proposed adaptive grid division approach. 
In uniform grid settings, the entire image is partitioned equally without considering object location. 
In contrast, our adaptive method performs fine-grained segmentation around the microrobot region and coarse segmentation in background areas, thereby allocating computational resources more effectively to informative regions.

As shown in Fig.~\ref{fig-ablation}, both mean squared error (MSE) and coefficient of determination ($R^2$) were used to evaluate the depth regression performance. 
Among the uniform grid configurations, the 4×4 setting achieved the best balance with an MSE of 0.034 and $R^2$ of 0.998. 
However, our adaptive grid strategy further outperformed all uniform divisions, yielding the lowest MSE of 0.027 and the highest $R^2$ of 0.999. 
These results confirm that incorporating structural priors via adaptive grids can significantly enhance the precision and generalisation of depth estimation.

\subsection{Influence of Data Size}
\begin{figure}[!t]
\centering
\includegraphics[width=1.0\hsize]{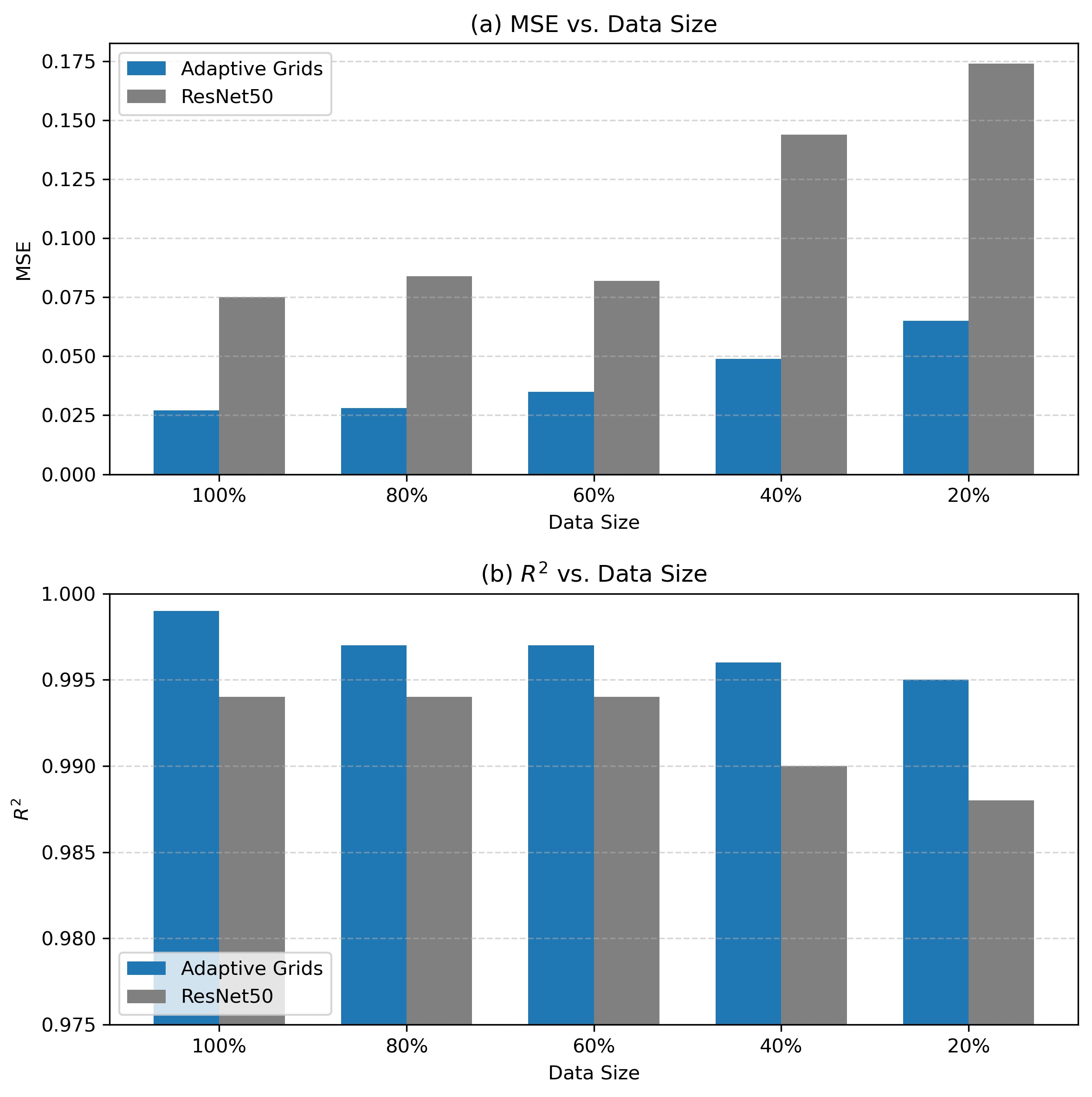}
\vspace{-0.7cm}
\caption{Effect of training-set size (Robot~1) on depth-regression performance.
(a) Mean squared error (MSE) and (b) coefficient of determination (\(R^2\)) for data availability levels of 100\%, 80\%, 60\%, 40\%, and 20\%.}
\label{fig-data_size}
\vspace{-0.65cm}
\end{figure}%

To evaluate the robustness of our method under limited data scenarios, we conducted an experiment using Robot 1 as a representative case. 
The full training set consists of 5,113 samples, and we trained models using progressively reduced subsets of the data: 100\%, 80\%, 60\%, 40\%, and 20\%, as shown in Fig.~\ref{fig-data_size}.

The results reveal three key findings:
First, as expected, the prediction accuracy decreases as the training data size becomes smaller. 
Both MSE increases and $R^2$ decreases consistently across models when less data is used.
Second, our method, based on physics-informed features and adaptive grid division, demonstrates significantly better resilience to data reduction compared to the ResNet50 baseline. 
Specifically, the increase in MSE from using 100\% to 20\% of the data is only 0.038 for our method, while the corresponding increase for ResNet50 is 0.099. 
This highlights that our approach is less reliant on large datasets and maintains stable performance even under data-constrained conditions.
Third, even when trained on only 20\% of the data, our method still outperforms ResNet50 trained with the full dataset in terms of both MSE and $R^2$. 
This result underscores the effectiveness of incorporating physical priors and adaptive spatial reasoning in achieving data-efficient depth regression for optical microrobots.

\section{Discussion}
The experimental results demonstrate the effectiveness of our proposed physics-informed depth estimation framework with adaptive grid-based feature extraction. 
Several factors contribute to this improved performance.
The adaptive grid strategy allows the model to allocate computational resources more efficiently by focusing on regions that are most informative, specifically, the microrobot itself. 
Traditional fixed-grid methods process all regions of the image equally, regardless of content. 
However, in our case, the background provides little to no information related to depth. 
By using finer grids in regions containing the microrobot and coarser grids elsewhere, our model captures detailed spatial variation where it matters most while avoiding unnecessary computation in uninformative areas. 
This not only improves computational efficiency but also enhances feature quality for depth regression.

The physics-based focus metrics, such as entropy, Laplacian of Gaussian, Tenengrad, Brenner measure, grey level variance, and maximum absolute gradient, were chosen based on their complementary characteristics in capturing focus and sharpness. 
These metrics respond differently to changes in image clarity caused by microrobot displacement along the optical axis. Combining them provides a richer, multidimensional representation of physical cues associated with depth. 
Compared to conventional, non-physically informed learning approaches, such as CNNs, VGG, ResNet, EfficientNet, and the transformer-based model ViT, as demonstrated in~\cite{wei2025dataset}, our proposed physics-informed method consistently outperforms all benchmarks. 
The key advantage lies in its ability to embed domain knowledge into the learning process, thereby enhancing the model’s sensitivity to depth variations and improving generalisation with limited data.

However, the current implementation relies on manually defined thresholds for microrobot detection and fixed grid sizes. 
To enhance adaptability and scalability, future work could explore automated and dynamic grid configuration strategies. For instance, instead of using static thresholds, learnable mechanisms such as attention-based models or transformer-guided frameworks could be employed to dynamically adjust grid resolution based on the spatial distribution of informative features.



\section{Conclusion}
In this work, we present a physics-informed and data-efficient framework for depth estimation of optical microrobots under monocular microscopic imaging. 
By incorporating physics-based focus metrics and designing an adaptive grid strategy that allocates computational attention according to microrobot location, our method enhances the estimation accuracy. 
Extensive experiments across multiple microrobot types and training data scales demonstrate the superior performance and robustness of our approach, particularly in low-data settings. 
Compared with standard CNN and ResNet-based models, our method achieves significantly lower MSE and higher $R^2$, even when trained on only a fraction of the available data. 
The proposed framework paves the way for more accurate and reliable 3D perception in microrobotic systems, offering valuable support for real-time, closed-loop micromanipulation tasks in biomedical applications.

\section*{Acknowledgement}
Lan Wei acknowledges support from a PhD studentship jointly sponsored by the China Scholarship Council and the Department of Bioengineering, Imperial College London.

\bibliographystyle{IEEEtran}
\bibliography{ref}

\end{document}